\documentclass{article}

\usepackage{arxiv}

\usepackage[utf8]{inputenc} % allow utf-8 input
\usepackage[T1]{fontenc}    % use 8-bit T1 fonts
\usepackage{hyperref}       % hyperlinks
\usepackage{url}            % simple URL typesetting
\usepackage{booktabs}       % professional-quality tables
\usepackage{amsfonts}       % blackboard math symbols
\usepackage{nicefrac}       % compact symbols for 1/2, etc.
\usepackage{microtype}      % microtypography
\usepackage{lipsum}		% Can be removed after putting your text content
\usepackage{graphicx}
\usepackage{natbib}
\usepackage{doi}

\title{PicArrange - Visually Sort, Search, and Explore Private Images on a Mac Computer}

\author{ 
    \href{https://orcid.org/0000-0002-3600-6848}{\includegraphics[scale=0.06]{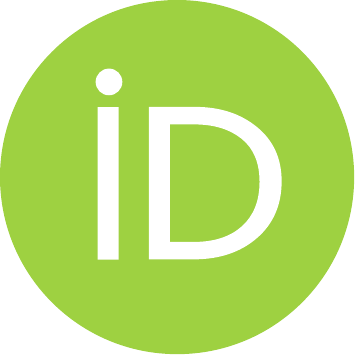}\hspace{1mm}Klaus Jung}\thanks{\url{https://visual-computing.com}} \\
    HTW Berlin, University of Applied Sciences\\ Visual Computing Group \\
    Wilhelminenhofstraße 75, \\ 12459 Berlin, Germany \\
    \texttt{klaus.jung@htw-berlin.de} \\
	\And
    \href{https://orcid.org/0000-0001-6309-572X}{\includegraphics[scale=0.06]{orcid.pdf}\hspace{1mm}Kai Uwe Barthel}\footnotemark[1] \\
    HTW Berlin, University of Applied Sciences\\ Visual Computing Group \\
    Wilhelminenhofstraße 75, \\ 12459 Berlin, Germany \\
    \texttt{kai-uwe.barthel@htw-berlin.de} \\
	\And
    \href{https://orcid.org/0000-0002-3957-4672}{\includegraphics[scale=0.06]{orcid.pdf}\hspace{1mm}Nico Hezel}\footnotemark[1] \\
    HTW Berlin, University of Applied Sciences\\ Visual Computing Group \\
    Wilhelminenhofstraße 75, \\ 12459 Berlin, Germany \\
    \texttt{nico.hezel@htw-berlin.de} \\
	\And
    \href{https://orcid.org/0000-0003-3548-0537}{\includegraphics[scale=0.06]{orcid.pdf}\hspace{1mm}Konstantin Schall}\footnotemark[1] \\
    HTW Berlin, University of Applied Sciences\\ Visual Computing Group \\
    Wilhelminenhofstraße 75, \\ 12459 Berlin, Germany \\
    \texttt{konstantin.schall@htw-berlin.de} \\
}

\renewcommand{\shorttitle}{PicArrange - Visually Sort, Search, and Explore Private Images}

\hypersetup{
pdftitle={PicArrange - Visually Sort, Search, and Explore Private Images on a Mac Computer},
pdfsubject={cs.CV, cs.LG},
pdfauthor={Klaus Jung, Kai Uwe Barthel, Nico Hezel, Konstantin Schall},
pdfkeywords={Content-based Image Retrieval, Exploration, Image Browsing,  
Visualization, CNNs, Image Sorting},
}

\begin{document}
\maketitle

\begin{abstract}
The native macOS application PicArrange integrates state-of-the-art image sorting and similarity search to enable users to get a better overview of their images. Many file and image management features have been added to make it a tool  that addresses a full image management workflow. A modification of the Self Sorting Map algorithm enables a list-like image arrangement without loosing the visual sorting. Efficient calculation and storage of visual features as well as the use of many macOS APIs result in an application that is fluid to use.%
%
%\footnote{The final authenticated publication is available online at \url{https://doi.org/[insert DOI]}}
\end{abstract}

% keywords can be removed

\section{Introduction}
\begin{figure}
    \centerline{\includegraphics[width=\textwidth]{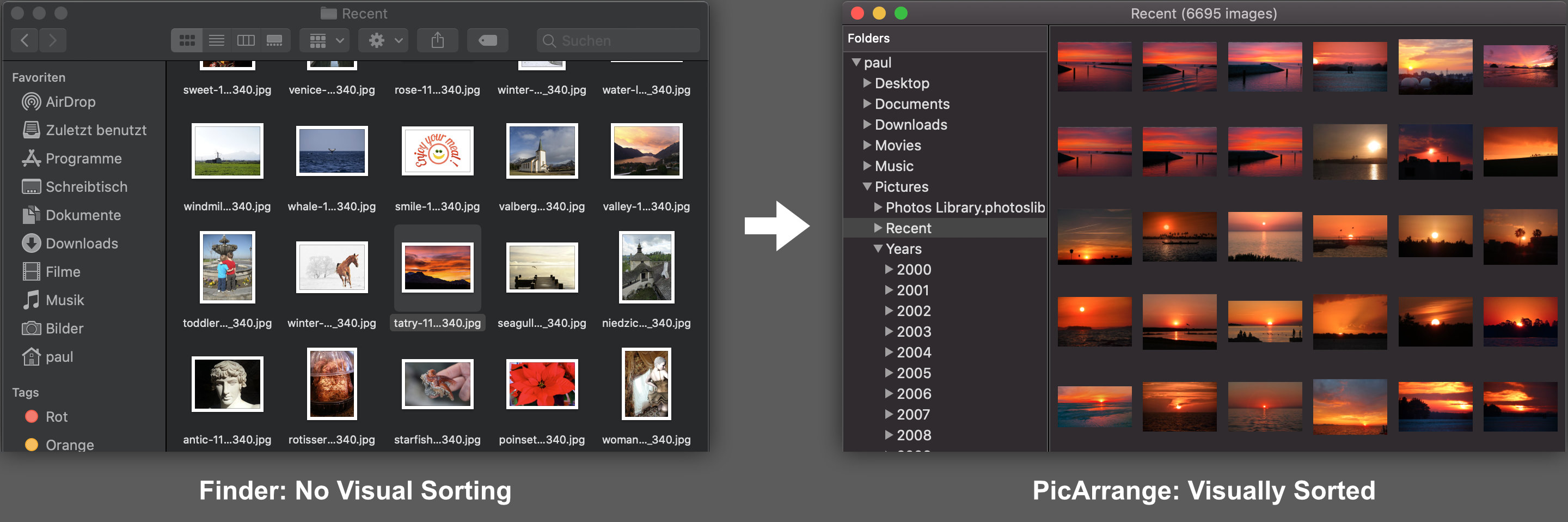}}
    \caption{Exploring a folder with 6695 images using macOS Finder and PicArrange}
    \label{fig:compare}
\end{figure}
\label{section:Introduction}
In many situations, people need to view and navigate through large amounts of images. This includes searching for appropriate images on the Internet or finding images from stock agencies. Visual similarity search helps to identify images that meet the search requirements. In most cases such systems are used to access images from a database using a web browser. However, with high-quality cameras built into smartphones the number of personal images keeps increasing, even if people are not professional photographers. Typically images are arranged as thumbnails sorted by file name or creation date. Surprisingly, visual image search is not integrated into common computer operating systems. Moreover, image lists only show images contained in a single folder of the file system. Fig. \ref{fig:compare} compares the display of images using the macOS' Finder and the proposed PicArrange application.

This paper introduces PicArrange, a native application highly integrated into the macOS to seamlessly work with images on the computer's file system. Here "file system" includes attached files storage as well as network shares or cloud drives. However, the focus is put to deal with private images without transferring any image data to a server for analysis purpose. PicArrange can display visually sorted images from several folders at the same time, making it easy to find duplicate images or finding the best looking image from a certain category. All analysis is done locally by calculating and storing the visual feature vectors for the displayed images to keep data privacy. By compressing the feature vectors, the amount of stored analysis data is reduced to a minimum, enabling PicArrange to handle hundreds of thousands of images simultaneously.
\section{Related Work}
\label{section:RelatedWork}
Many web-based image management systems offer the possibility to find "similar images", such as Google Images\footnote{\url{https://images.google.com}}%
, Pixabay\footnote{\url{https://pixabay.com}}, Pexels\footnote{\url{https://www.pexels.com}}, Adobe Stock\footnote{\url{https://stock.adobe.com}}, and others. In many cases, similarity search is based on tagged keywords, some systems use visual comparison or a combination of both. In most cases, the  algorithms used are not published. A few systems use a visually sorted arrangement of the presented images. Wikiview\footnote{\url{https://wikiview.net}} \cite{wikiview} uses Self-Sorting Maps (SSM) \cite{Strong:2014:SME:2719262.2719635} for visual sorting to allow browsing of Wikimedia Common images. Picsbuffet\footnote{\url{https://picsbuffet.com/pixabay}} \cite{10.1007/978-3-319-14442-9_30} can be used to explore images from Pixabay, Fotolia, and IKEA. 

Operating systems like macOS, iOS, or Windows nowadays do some analysis on local images using machine learning approaches for image classification and identification. On macOS and iOS, the images need to be part of Apple's Photo Library to be analysed. Most visibly, faces found in the images are related to certain persons in a semi-automatic approach. Depending on the amount of images, the analysis takes hours or days. It does not work for images from the file system that have not been imported into the Photo Library. Kiano\footnote{\url{https://visual-computing.com/project/kiano}} uses Apple's Photo Library API to display images visually arranged similar to Picsbuffet. ImageX\footnote{\url{https://visual-computing.com/project/imagex}} \cite{10.1145/3206025.3206093} is a multi-platform, Java-based application that allows to visually explore and search images from the file system of Linux, Windows, and Mac operating systems.
\section{PicArrange}
\label{section:PicArrange}
The presented demo application PicArrange differs from the applications mentioned in Section \ref{section:RelatedWork} in several aspects. It is neither web-based nor does it rely on a web service for image analysis. Compared to ImageX, it is not platform independent, which might be a disadvantage at first sight. However, PicArrange is designed to integrate into macOS as much as possible, offering features more than just image sorting (see Section \ref{subsection:Integration}), thus addressing users that need a complete workflow in image management and want to retain to the look and feel of a native application, using interaction methods to which they are already accustomed. Performance, i.e. fast processing times, is also a requirement that led to the decision to use native APIs as much as possible.

\begin{figure}%[h]
    \begin{tabular*}{\linewidth}{@{\hspace{0em}}c@{\extracolsep{\fill}}c@{\hspace{0em}}}
    \includegraphics[height=0.4\linewidth]{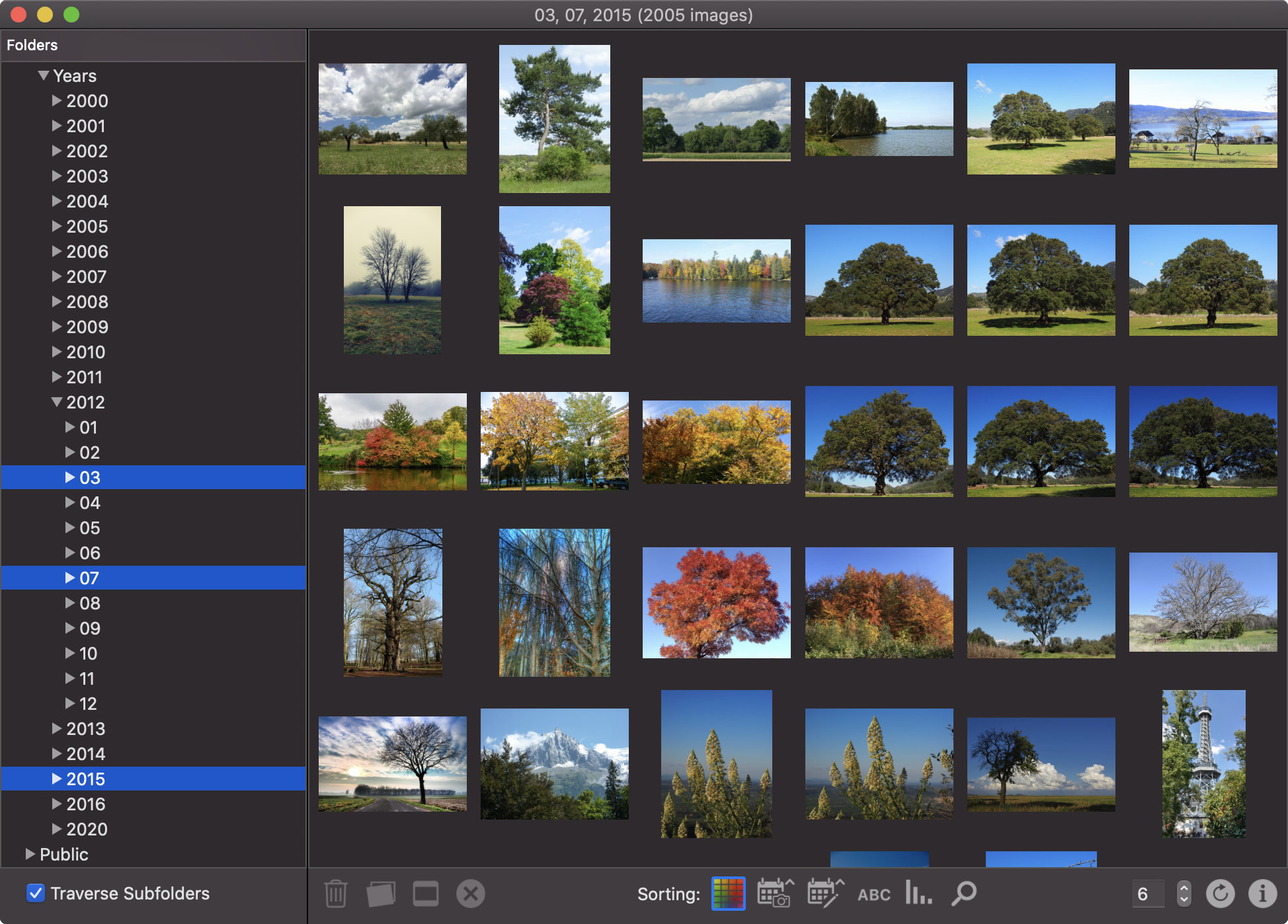} &
    \includegraphics[height=0.4\linewidth]{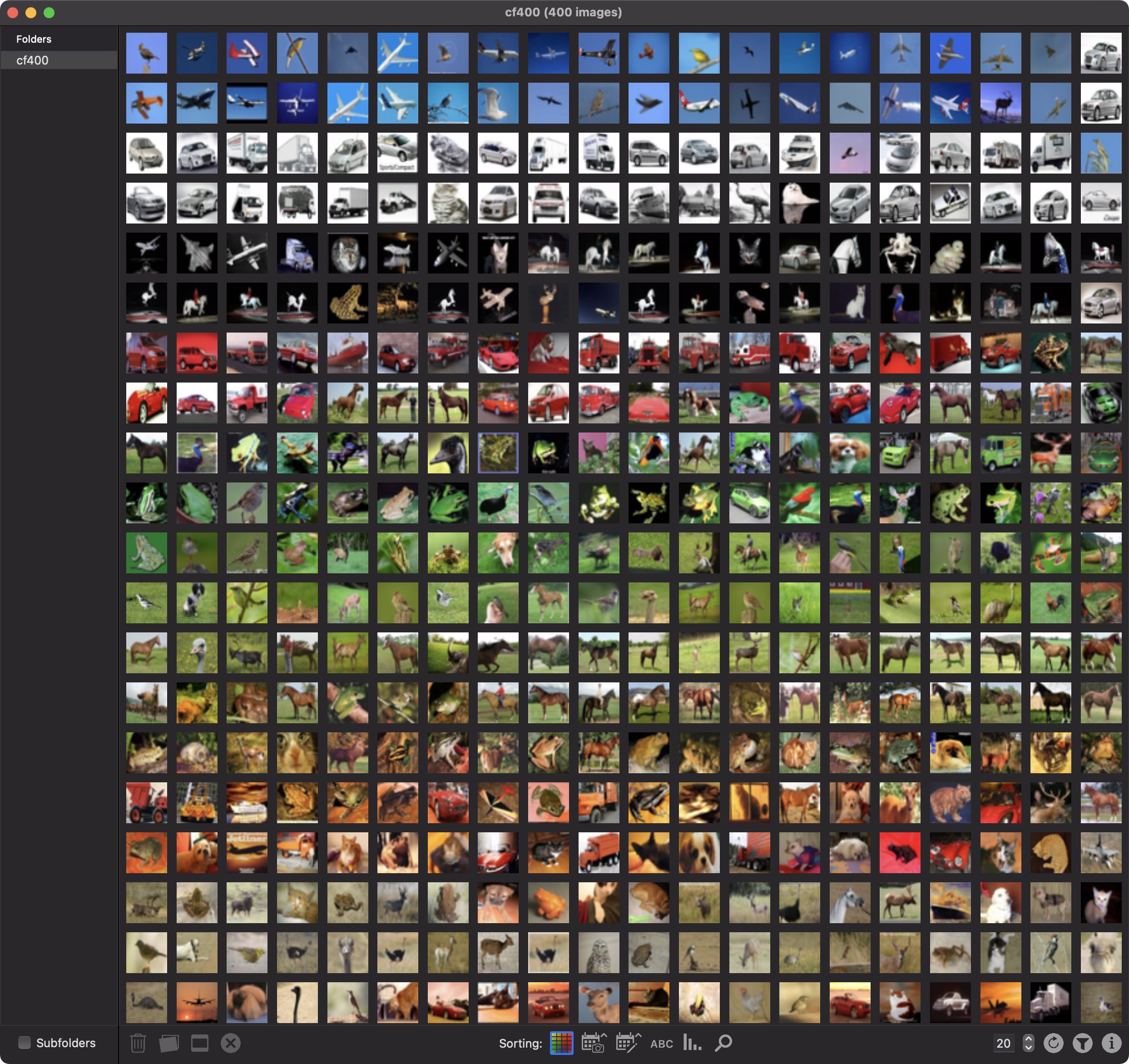} \\
    (a) & (b)
    \end{tabular*}
    \centering
    \caption{Viewing images from multiple folders (a) and with many columns (b)}
    \label{fig:multipleFolders}
\end{figure}

The most visible difference compared to ImageX, Kiano, or Picsbuffet is the arrangement of sorted images. Whereas these applications display an endless map of visually sorted images that topologically represents a torus, PicArrange uses a "linear" column-based layout (Fig. \ref{fig:multipleFolders}). 
This adopts the presentation style of the Finder application. A map that is almost square in size and repeats endlessly at the edges is a good approach to get an overview of a huge amount of images.  Visual sorting benefits from many places in both dimensions to arrange similar images nearby. But in most cases, exploring private images on the local file system the users do not display thousands of images at once. A list of images with a small number of columns ensures that the user knows when he or she has seen all images in that folder because the list simply has an end. Nevertheless, visually sorting images in this layout leads to an improved visualization (see Section \ref{subsection:ArrangingImages} and Fig. \ref{fig:compare}).

Unlike the Finder, PicArrange can display images from multiple folders at once (Fig. \ref{fig:multipleFolders} (a)). 
It is also possible to include images from the subfolders of a folder, so that all images of the subtree of this folder are displayed. This allows the user to get an overview of the images, even when they are arranged in a complex folder structure.

A short introduction to PicArrange with many screenshots can be found at \url{https://visual-computing.com/project/picarrange/help}. The application is available in the App Store\footnote{Download PicArrange from \url{https://apps.apple.com/app/picarrange/id1530678223}}.

\subsection{Feature Calculation}
\label{subsection:Feature}
For a good compromise between processing time, sorting quality, and storage size of feature vectors, PicArrange computes features using the MobileNetV3 CNN \cite{conf/iccv/HowardPALSCWCTC19}.
The pretrained network is modified in such a way that activations before the fully connected layer are passed to a compression network integrated to the model. This network reduces the data to 64 dimensions and can be seen as an alternative to Principal Component Analysis. Converted to Apple's CoreML API, image features can be calculated in a single step using the GPU processing power of the Mac computer.

When displaying only small thumbnails, details of the image content cannot be perceived. The subjective quality of a visually sorted image arrangement is increasingly influenced by the colors of the images compared to their content (compare Fig. \ref{fig:multipleFolders} (a) with (b)). For this reason, we calculate a second feature vector not using a neural network. It calculates its values from color histograms, edge histograms, and a frequency analysis. Finally, a weighted combination of CNN and non-CNN features is used, with CNN features given a higher weight in PicArrange's similarity search (content is more important than color) compared to the visually sorted image arrangement. The optimal weighting for the visual search was determined using mean average precision calculations, and for visual sorting the weighting was adjusted to provide a good subjective viewing experience of the sorted images.  

\subsection{Image Arrangement}
\label{subsection:ArrangingImages}
Self-Organizing Maps (SOM) or Self-Sorting Maps (SSM) arrange images on a grid of $N \times M$ positions. Within these algorithms, border processing at the grid's boundaries is needed. When creating a seamless map of endless repeated images on a torus, calculations in the vicinity of the borders need to wrap around grid positions. The last images of a row is followed by the first image of that row. For PicArrange's list with a few columns this would require that the last images of a row need to be similar to the first one. Such a restriction is not necessary and would make it more difficult for the algorithm to find good places for the images in such a narrow grid. Thus, a border processing of constant continuation is used within PicArrange's image sorting process.

\begin{figure}[t]
    % Dieses Bild musste ich ziemlich klein machen, damit der Platz reicht
    \centerline{\includegraphics[width=0.2\textwidth]{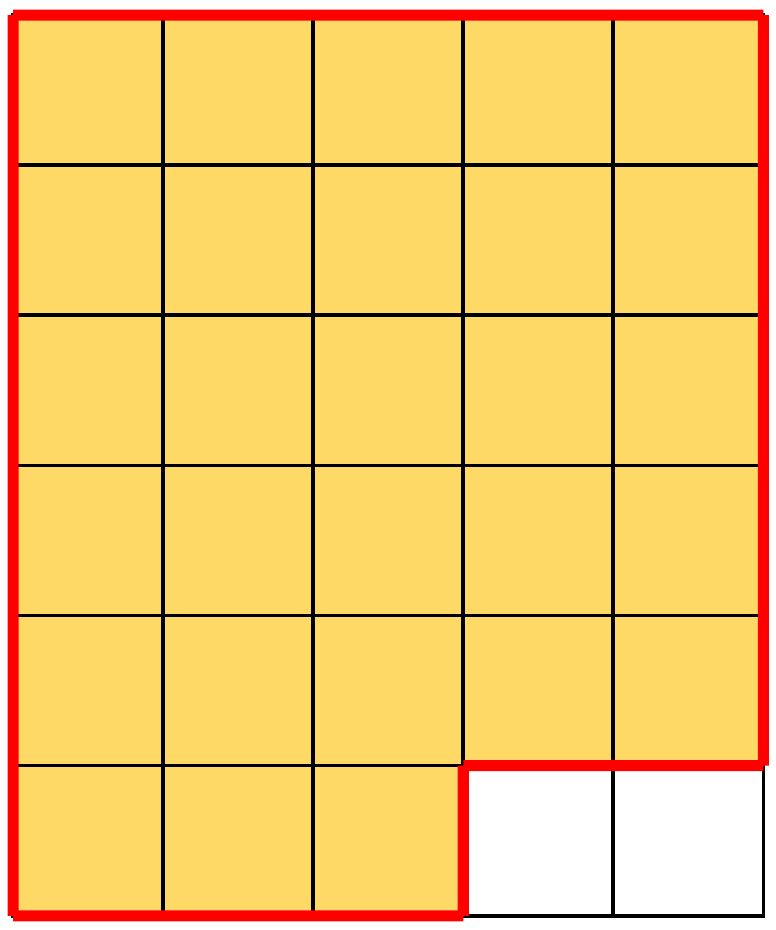}}
    \caption{Border (red) of 28 images placed on $5 \times 6$ positions}
    \label{fig:border}
\end{figure}

Most applications using SOM or SSM show holes or duplicate images at certain map positions if the number of images is not a multiple of the number of columns $N$. Holes can be easily created by padding the unsorted array with duplicate images carrying a flag "do not show in final result". However, the holes will be at "random" positions of the sorted array. An application used to explore images in the file system should not show duplicates unless there are duplicate files. So both options do not meet the user's expectation of a list without duplicates and without holes, expect at the last positions of the last row. To achieve this goal, PicArrange uses a modified constant continuation border processing where the border is not simply a rectangle, but the shape around images filling up $N \times M$ positions in scanline order (Fig. \ref{fig:border}).
\subsection{Similarity Search}
\label{subsection:Similarity}
Similarity search is implemented by selecting a query image and sorting all images of the list by their feature vector's L2-distance to the feature vector of the query image. To improve the usability of the application, other things can also be done. Searching for similar images can be an iterative process where the user identifies images from the first result that fit the expectation even better than the query image. Often there is the need to use more than just one image for the query. \url{https://visual-computing.com/project/picarrange/help/?p=12} shows screenshots for an iterative search looking for windmills and towers. When multiple images are used as query, the result images are sorted by the smallest distance to one of the query images.

PicArrange allows to select query images from a certain folder and to display search results taken from any extended folder selection, e.g. a super-folder of the query folder. This is a typical situation when the user knows where the query image is located in the file system, but wants to find similar images throughout the entire collection. PicArrange always tries to keep the user selection of images, even when changing folders.
\subsection{Integration Into macOS}
\label{subsection:Integration}
The main features apart from image sorting and similarity search are listed below. They have been implemented by using many of Apple's APIs.
\begin{itemize}
    \item Use drag \& drop into the Finder to copy images to any other folder.
    \item Delete selected image files.
    \item Show image files in Finder or open them with macOS' Preview application.
    \item Open images with any macOS application registered to view or edit images.
    \item Sort images by creation date, modification date, file name, or file size.
    \item View images from multiple folders at once.
    \item Display images of all macOS-supported image file formats, including camera manufacturers' RAW formats.
    \item Display videos and first pages of PDF documents as well.
    \item Display image file information.
    \item Support for attached storage devices, network shares, and iCloud folders.
    \item Filter displayed images by criteria like file name, creation date, modification date, or size.
    \item PicArrange also has a build-in single image viewer with zoom, full screen and slideshow.
\end{itemize}
Processing time depends on the time used for traversing the file system, acquiring thumbnails from the OS, feature calculation, and image sorting. For a set of 5300 typical digital camera images (JPEG) on the local SSD, the total processing time is about 340 seconds (64 ms per image) including feature calculation (Intel i7-6820HQ, Radeon Pro 455 2 GB). For a second run on the same set, PicArrange uses its stored feature vectors. The processing time reduces to 1.7 seconds (0.32 ms per image).

\section{Conclusion}
\label{section:Conclusion}
This demo presents a highly integrated viewer for images and videos on a macOS file system that allows users to explore and process all their images. Using CNN-based image features and optimized image sorting, users benefit from improved visualization and image search without sacrificing familiar functionality.

\bibliographystyle{unsrtnat}
\bibliography{references}  %%% Uncomment this line and comment out the ``thebibliography'' section below to use the external .bib file (using bibtex) .

\begin{thebibliography}{5}
\providecommand{\natexlab}[1]{#1}
\providecommand{\url}[1]{\texttt{#1}}
\expandafter\ifx\csname urlstyle\endcsname\relax
  \providecommand{\doi}[1]{doi: #1}\else
  \providecommand{\doi}{doi: \begingroup \urlstyle{rm}\Url}\fi

\bibitem[Barthel et~al.(2019)Barthel, Hezel, Schall, and Jung]{wikiview}
Kai~Uwe Barthel, Nico Hezel, Konstantin Schall, and Klaus Jung.
\newblock Real-time visual navigation in huge image sets using similarity
  graphs.
\newblock In \emph{Proceedings of the 27th ACM International Conference on
  Multimedia}, MM '19, page 2202–2204, New York, NY, USA, 2019. Association
  for Computing Machinery.
\newblock ISBN 9781450368896.
\newblock \doi{10.1145/3343031.3350599}.

\bibitem[Strong and Gong(2014)]{Strong:2014:SME:2719262.2719635}
Grant Strong and Minglun Gong.
\newblock Self-sorting map: An efficient algorithm for presenting multimedia
  data in structured layouts.
\newblock \emph{Trans. Multi.}, 16\penalty0 (4):\penalty0 1045--1058, June
  2014.
\newblock ISSN 1520-9210.
\newblock \doi{10.1109/TMM.2014.2306183}.

\bibitem[Barthel et~al.(2015)Barthel, Hezel, and
  Mackowiak]{10.1007/978-3-319-14442-9_30}
Kai~Uwe Barthel, Nico Hezel, and Radek Mackowiak.
\newblock Imagemap - visually browsing millions of images.
\newblock In Xiangjian He, Suhuai Luo, Dacheng Tao, Changsheng Xu, Jie Yang,
  and Muhammad~Abul Hasan, editors, \emph{MultiMedia Modeling}, pages 287--290,
  Cham, 2015. Springer International Publishing.
\newblock ISBN 978-3-319-14442-9.

\bibitem[Hezel and Barthel(2018)]{10.1145/3206025.3206093}
Nico Hezel and Kai~Uwe Barthel.
\newblock Dynamic construction and manipulation of hierarchical quartic image
  graphs.
\newblock In \emph{Proceedings of the 2018 ACM on International Conference on
  Multimedia Retrieval}, ICMR '18, page 513–516, New York, NY, USA, 2018.
  Association for Computing Machinery.
\newblock ISBN 9781450350464.
\newblock \doi{10.1145/3206025.3206093}.

\bibitem[Howard et~al.(2019)Howard, Pang, Adam, Le, Sandler, Chen, Wang, Chen,
  Tan, Chu, Vasudevan, and Zhu]{conf/iccv/HowardPALSCWCTC19}
Andrew Howard, Ruoming Pang, Hartwig Adam, Quoc~V. Le, Mark Sandler, Bo~Chen,
  Weijun Wang, Liang-Chieh Chen, Mingxing Tan, Grace Chu, Vijay Vasudevan, and
  Yukun Zhu.
\newblock Searching for mobilenetv3.
\newblock In \emph{ICCV}, pages 1314--1324. IEEE, 2019.
\newblock ISBN 978-1-7281-4803-8.
\newblock URL
  \url{http://dblp.uni-trier.de/db/conf/iccv/iccv2019.html#HowardPALSCWCTC19}.

\end{thebibliography}





% \end{thebibliography}

\end{document}